# Devanagari Handwritten Character Recognition Using TrOCR: A Transformer-Based Model with Real-Time Web Deployment

Amrit Baskota
*School of Computer Science and Engineering*
*Vellore Institute of Technology*
Vellore – 632014, Tamil Nadu, India
amrit.baskota2022@vitstudent.ac.in

Samyam Budhathoki
*School of Computer Science and Engineering*
*Vellore Institute of Technology*
Vellore – 632014, Tamil Nadu, India
samyam.budhathoki2022@vitstudent.ac.in

Shubham Ghimire
*School of Computer Science and Engineering*
*Vellore Institute of Technology*
Vellore – 632014, Tamil Nadu, India
shubham.ghimire2022@vitstudent.ac.in

Abiskar Ghimire
*School of Computer Science and Engineering*
*Vellore Institute of Technology*
Vellore – 632014, Tamil Nadu, India
abiskar.ghimire2022@vitstudent.ac.in

Sarwesh Phuyal
*School of Computer Science and Engineering*
*Vellore Institute of Technology*
Vellore – 632014, Tamil Nadu, India
sarwesh.phuyal2022@vitstudent.ac.in

Baskaran P *
*School of Computer Science and Engineering*
*Vellore Institute of Technology*
Vellore – 632014, Tamil Nadu, India
baskaran.p@vit.ac.in

*Abstract*— **Devnagari is a one of the ancient language of the Indian subcontinent consisting of 36 vowels, 14 consonants and 10 numerals. The accurate recognition of handwritten Devnagari characters is challenging due to high complexity of Devnagari scripts. This paper presents a method to fine tune the pre trained TrOCR model to accurately recognize Devanagari handwritten characters. The Methodology consists of a preprocessing mechanism where input images are standardized into RGB format, tokenized in batches and integrated with Hugging Face Dataset. The pre-trained microsoft/trocr-base-handwritten model is fine-tuned over a dataset of nearly 5000 character images that are uniformly partitioned in the ratio 8:1:1 for training, evaluation and testing. Further optimization is done is the training process through mixed precision training, gradient checkpointing, and early stopping mechanism. The model achieves a character error rate (CER) of 3.95% and a character-level accuracy of 96.05%, outperforming the previous CNN based models. A scalable web application is developed using Vite.js, Golang, and FastAPI which practically deploys the OCR model and serves character recognition task with a latency less than 5 seconds per request. This study demonstrates the use of TrOCR model to build a scalable handwritten Devnagari character recognition system and also a foundation to future research on Devnagari Script Recognition using Transformers.**

***Keywords—Devnagari Handwritten Character Recognition (DHCR), TrOCR, RoBERTa, CNN, CER, Golang, Fast-API, Fine-Tuning***

## I. Introduction

In India and Nepal both Hindi and Nepali are widely spoken language. Both languages are derived from one of the oldest languages of world i.e. Sanskrit. Having the same root, both languages use Devanagari Script while writing. The Devanagari script is written in a very complex way combining various symbols, consonants, vowels and symbolic marks which forms a highly structured writing system which can make Optical Character Recognition (OCR) challenging as it must accurately detect and process detailed character combinations, accent marks and positional change for accurate text extraction.

An OCR is a system which translates the images which it takes by reading any text from paper into a form that computer can manipulate. In recent years OCR has gained a major importance because of its unique features. As OCR focuses in greater efficiency by depending upon the nature of script, it's development is the area of interest for several researchers.

Several OCR systems are already created for recognising English language followed by other European languages. The OCR system for Devanagari script is still on it's developing stage and a lot of work is needed to be done for making this useful. This paper presents a system for detecting Devanagari script using the latest TrOCR model, improving upon traditional CNN approaches. Our main goal is to compare the accuracy of this modern technique with the traditional one based on key factors such as character recognition efficiency, processing speed, and error rate. By evaluating these metrics on a dataset of handwritten Devanagari text, we aim to demonstrate the advantages of TrOCR in handling complex character combinations, accent marks and positional changes more effectively than CNN based methods.

## II. Related Work

Several attempts have been made in recent years to develop text recognition systems for various languages and scripts. OCR research has been ongoing for over three decades with conventional algorithms such as Support Vector Machine (SVM) and Hidden Markov Model (HMM) being widely used. However, these methods rely on explicit character-level segmentation, which is challenging to interpret and limits their effectiveness in recognizing complex scripts like Devanagari. The noise suppression and accurate character recognition are challenged by the variation of the styles of the handwritten text. Because of such challenges ensuring a high accuracy in the script recognition remains a difficult task.

The early research on Devanagari Scripts dates back to 1977. Though it achieved a limited success, it has been a source through which the further attempts are made to refine the system. The script detection is hard because of the joining of characters, not only joining sometimes the characters are merged into compound characters. Devanagari scripts consists of ”matras” (diacritical marks). These ”matras” modifies the fundamental vowel sounds of the consonants which also result in the change of the pronunciation and sometimes meaning too which makes the character recognition more complex.

Neha Sahu and Nitin Kali Raman [2] developed a Devnagari character recognition system that assembles the Artificial Neural Networks for pre-processing, segmentation and recognition. Even though the character recognition of Devnagari script is very challenging, the system developed was tested on the noisy characters and achieved an accuracy of 75.6%.

Shunya Rakuka, Kento Morita and Tetushi Wakabayashi [4], on the other hand, focused on the handwritten character string recognition and showcased a Transformer-CNN-based method. In this method, there is no requirement of a large annotated dataset and also achieved a CER of 0.241. Later they incorporated a Transformer Decoder and type string images through which there was a significant improvement in the accuracy of the system by reducing the CER to 0.164 and further to 0.127 after adding the training data.

There are further advancements that have been made in OCR which include the integration of CNN-Transformer architectures for many languages, ensuring improved accuracy [1]. The application of deep learning techniques, particularly CNN and RNN architectures, has led to significant improvements in handling the complexity and variability of handwritten Devanagari script. Hybrid methods that integrate structural and statistical techniques have further enhanced recognition accuracy while maintaining computational efficiency [7]. Moreover, the development of well-annotated datasets has helped a lot to strengthen the training and evaluation processes which has led to more reliable OCR systems.

When ResNet-based features extractors is combined with data augmentation techniques like GANs the generalization across diverse handwriting styles is enhanced. Similarly, the layout analysis techniques using multi-stage localization with DoG and SIFT descriptors have provided great benefit to the historical manuscript OCR. Clustering methods like DBSCAN and keyword spotting BAG enhance the text-line segmentation and character recognition [3]. Moreover, PCA-based ruling evaluation refines the text attraction in degraded manuscripts along with semi-supervised learning through hybrid approaches like GIDOC. By combining the deep learning-based feature extraction and the traditional layout analysis methods the recognition performance across the diverse scripts can further be enhanced.

[6] The paper explores a range of deep models like CNN-BiLSTM and transfer learning algorithms like VGG-16 and ResNet-50 to enhance recognition accuracy. The paper also focuses to enhance generalization by using synthetic data augmentation in models. All these enhancements address handwritten OCR problems like character segmentation. Recent developments in OCR have been aimed at translating handwritten Devanagari text into editable text using CNN-based architectures for improving recognition accuracy [8]. The work has addressed issues of character merging, half consonants, and vowel modifiers. These are a few of the key challenges of text recognition. CNNs do the classification and feature extraction. This improves the accuracy of OCR systems. By integrating CNN-based methods with robust pre-processing, OCR engines would be better equipped to handle handwriting style variations.

[9] Complementing these advancements, word detection and localization methods utilizing morphological image processing have refined segmentation techniques for handwritten Devanagari text. These approaches effectively address challenges such as overlapping and touching characters by improving character boundary definition. Moreover, advanced pre- processing techniques have contributed to better feature extraction, ensuring higher accuracy in handwritten OCR applications.

[12] Machine Learning models like CRNN, SeqCLR, Vision Transformers (ViT), and TrOCR were used in renAIssance project to recognize historical documents consisting of advanced formatting and font variations. The TrOCR model outperformed every other models in terms of accuracy with a CER of 0.024). This indicates that Transformer-based architectures are capable of addressing issues such as inconsistent text formats and ambiguous characters.

[13] The paper compares a CNN-GRU model with a fine-tuned TrOCR with self-attention model for handwritten text recognition. While TrOCR achieved the highest accuracy, the CNN-GRU model was able to provide a lightweight and competitive performance. This demonstrates that self-attention enhances recognition even during limited-resources.

[14] The paper presents a template matching-based approach for offline handwritten Devanagari character recognition. The system was able to achieve an accuracy of 92.66%, by the use of pre-processing and normalization techniques to handle character variability and noise in scanned images.

[15] This paper introduces an AI-based system to evaluate handwritten exam papers using a combination of OCR and NLP techniques. It uses models like TrOCR and EasyOCR for text extraction and uses cosine similarity and transformer-based methods for answer comparison. This showcases the real-life usage of the TrOCR model.

[16] This paper presents a comparative study of various machine learning and deep learning algorithms like SVM, MLP, and CNN for to recognize handwritten digits. CNN-based architectures were found efficient for classification tasks, and Dilated Temporal Convolution Networks (DTCN) were used for extracting temporal features, along with some scale-space techniques for effective word segmentation.

## III. Proposed Model

In this study, we propose a modern approach to text recognition for the Devanagari script using the TrOCR model. Even though there are many researches being done in OCR for decades, recognizing handwritten Devanagari remains challenging. The factors that make Devanagari OCR difficult are character merging, compound characters, and the presence of “matras.” Traditional methods such as SVM and HMM required explicit character segmentation. That segmentation limited their effectiveness for complex scripts. Researchers have used deep learning methods like CNNs, RNNs, and Transformers for a long duration to make text recognition more accurate. Recent studies have shown that combining models like CNN-Transformer, along with data augmentation techniques such as GANs, improves the system’s ability to handle different handwriting styles. These results from various existing systems are taken as references while designing our model. Despite recognizing handwritten and historical Devanagari text is very difficult because of less practice, TrOCR, a Transformer-based model, has demonstrated strong performance in the character detection. This model can do so because of the improvements like advanced pre-processing, layout analysis, and clustering methods which helped achieve better segmentation and character recognition. This system also proves itself that it is better than the traditional methods like SVM, HMM and CNN.

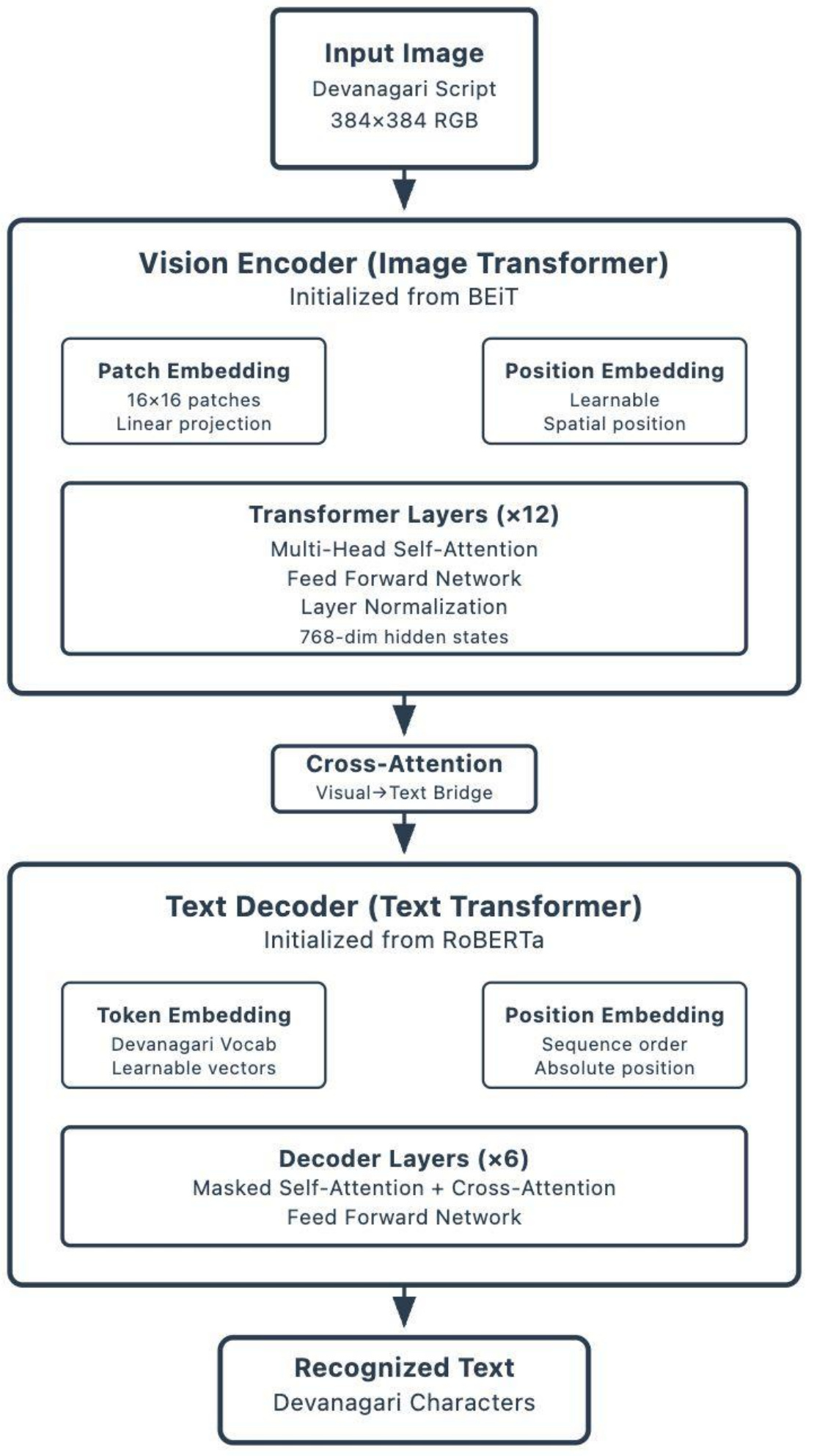

Fig. 1. Proposed TrOCR Model Architecture

The proposed TrOCR model processes handwritten images of Devnagari characters using a transformer-based encoder-decoder architecture. Input image (384×384) having handwritten Devanagari script is given to the system. In the Vision Encoder Components patch embedding, the 384×384 input image is converted into non-overlapping 16×16 patches, and then a linear projection is applied to produce patch tokens. It adds learnable spatial location data to each patch token to preserve spatial relationships. Transformer layers contain 12 sequential blocks, each containing multi-head self-attention mechanisms and feed-forward networks. The encoder gives 768-dimensional feature replicas for each image patch. The cross-attention bridge allows information flow between visual and textual modalities. The text decoder components use token embedding to convert Devanagari characters to dense vector representations, using a vocabulary matrix. Position embedding tracks character order in the text sequence. Masked attention makes sure that the model only uses past information when predicting the next character. Cross-attention connects the text decoder to visual features from the image encoder. Devanagari characters are generated successively as the final output.

## IV. Methodology

### A. Dataset Preprocessing

The dataset used for this research are handwritten images of Devnagari characters. The overall dataset is divided into 3 separate categories for Devanagari numerals, vowels and consonants. The class, label and phonetics of each character is maintained in a separate csv file for each category. The dataset is divided uniformly into train set, validation set and test set in a ratio 8:1:1.

```
Numerals Labels:
   Class  Label Devanagari Label Phonetics
0      0      0                ०    Śūn'ya
1      1      1                १       ēka
2      2      2                २      du'ī
3      3      3                ३      tīna
4      4      4                ४      cāra
Vowels Labels:
   Class Label Devanagari Label Phonetics
0      1     a               अ         a
1      2    aa               आ         ā
2      3     i               इ         i
3      4    ee               ई         ī
4      5     u               उ         u
Consonants Labels:
   Class Label Devanagari Label Phonetics
0      1    ka               क        ka
1      2   kha               ख       kha
2      3    ga               ग        ga
3      4   gha               घ       gha
4      5   kna               ङ        ṅa
```

Fig. 2. Labels

```
Train Set: 3748 images
Validation Set: 467 images
Test Set: 467 images
```

Fig. 3. Splitting

In order to make sure that consistency is maintained, preprocessing is done to each character image. Initially to make the dataset compatible with TrOCR, the *DataFrames* are converted into *Hugging Face Datasets* and images is converted to RGB format. Then images are converted to numerical pixel values whereas the labels are converted into tokenized IDs. The preprocessing is done in batches so that multiple samples can be processed at once to ensure efficiency.

### B. *Model Training and Fine-Tuning Hyperparameters*

A fine-tuned TrOCR model for Devnagari Character Recognition should be based on an encoder-decoder framework. As the pre-trained TrOCR base model, microsoft/trocr-base-handwritten is loaded from VisionEncoderDecoder class of the transformers library. This TrOCR model is an encoder-decoder model where an image Transformer is used as encoder, and a text Transformer is used as decoder.

1%▌ | 20/1404 [09:12<6:45:32, 17.58s/it]{'eval_loss': 0.9973840713500977, 'eval_runtime': 195.2451, 'eval_samples_per_second': 2.392, 'eval_steps_per_second': 0.302, 'epoch': 0.09}

3%▌ | 40/1404 [18:28<6:55:25, 18.27s/it]{'eval_loss': 0.8705172538757324, 'eval_runtime': 200.898, 'eval_samples_per_second': 2.325, 'eval_steps_per_second': 0.294, 'epoch': 0.17}

4%▌ | 60/1404 [27:20<6:32:51, 17.54s/it]{'eval_loss': 0.7835668921470642, 'eval_runtime': 184.1208, 'eval_samples_per_second': 2.536, 'eval_steps_per_second': 0.32, 'epoch': 0.26}

6%█ | 80/1404 [36:35<6:37:34, 18.02s/it]{'eval_loss': 0.7920494675636292, 'eval_runtime': 190.421, 'eval_samples_per_second': 2.452, 'eval_steps_per_second': 0.31, 'epoch': 0.34}

7%█ | 100/1404 [42:22<6:18:55, 17.44s/it]{'loss': 1.3843, 'grad_norm': 8.061577796936035, 'learning_rate': 1.8575498575498575e-05, 'epoch': 0.43}

==========================================

93%|█████████| 1300/1404 [9:32:42<29:49, 17.21s/it]{'eval_loss': 0.03223033621907234, 'eval_runtime': 190.3197, 'eval_samples_per_second': 2.454, 'eval_steps_per_second': 0.31, 'epoch': 5.55}

94%|█████████| 1320/1404 [9:41:33<24:05, 17.21s/it]{'eval_loss': 0.03163674473762512, 'eval_runtime': 188.2706, 'eval_samples_per_second': 2.48, 'eval_steps_per_second': 0.313, 'epoch': 5.64}

95%|█████████ | 1340/1404 [9:50:09<18:17, 17.15s/it]{'eval_loss': 0.03146235644817352, 'eval_runtime': 174.5873, 'eval_samples_per_second': 2.675, 'eval_steps_per_second': 0.338, 'epoch': 5.72}

97%|█████████| 1360/1404 [10:00:09<13:56, 19.01s/it]{'eval_loss': 0.03172135725617409, 'eval_runtime': 248.1926, 'eval_samples_per_second': 1.882, 'eval_steps_per_second': 0.238, 'epoch': 5.81}

98%|█████████| 1380/1404 [10:10:52<07:51, 19.65s/it]{'eval_loss': 0.0317654050886631, 'eval_runtime': 251.9398, 'eval_samples_per_second': 1.854, 'eval_steps_per_second': 0.234, 'epoch': 5.89}

100%|█████████| 1400/1404 [10:17:24<01:18, 19.65s/it]{'loss': 0.0009, 'grad_norm': 0.012886188924312592, 'learning_rate': 5.6980056980056986e-08, 'epoch': 5.98}och': 5.98}

Fig. 4. Training Log

The image encoder was initialized from the weights of BEiT, while the text decoder was initialized from the weights of RoBERTa. The raw model is initially capable for optical character recognition on images with single line handwritten English sentences.

To begin with fine-tuning process, training arguments are set. Two samples are trained per batch. Gradients are accumulated to the batch at every 4th step. The model is trained for maximum six epochs. Early stopping is enabled to prevent unnecessary training if no improvements are observed or if the model starts to overfit.

The evaluation strategy is set to "steps" and the model is evaluated after every 100 steps. Logging is done after every 100 steps to monitor the training progress. The model checkpoints are saved after every 500 steps to prevent any risk of data loss.

For further optimization, gradient checkpointing, mixed-precision training(fp16) are implemented.

The training process was steady. Initially the evaluation loss was observed to be 0.9974 at 1% of training completion i.e. 0.09 epochs where evaluation runtime was 195.24 seconds with 2.79 samples processed per second. The evaluation loss decreased to 0.1279 at 7% of training i.e. 1.11 epochs with 2.52 samples processed per second. This indicated that the evaluation loss was reducing gradually with an improvement in the evaluation efficiency after every step.

At the end of training process, evaluation loss reduced to approximately 0.031 with total evaluation runtime of 174.58 seconds. Additionally, the training loss was also observed to reduce from 1.3843 at the beginning of the training to around 0.0009 at the end.

### C. *Evaluation Metrics and Performance Analysis*

The model’s performance is analyzed based on its predictions over the test set. For evaluating model performance, Word Error Rate (WER) and Character Error Rate (CER) are used. The predictions made by model are compared to actual labels and WER and CER are computed as:

$$\text{CER} = \frac{S + D + I}{C}$$

S = Number of character substitutions

D = Number of deletions

I = Number of insertions

C = Total number of characters in the reference text

Suppose:

Reference: ”क” (1 character).

Prediction: ”ख” (1 character).

CER Calculation:

S=1 (substitution: क → ख).

D=0, I=0

C=1

$$\text{CER}=\frac{1+0+0}{1}=1.0\ (100\%)$$

```
100%|██████████| 59/59 [06:57<00:00,  7.07s/it]
{'CER': 0.039525691699604744, 'Char Accuracy': 0.9604743083003953}
```

Fig. 5. CER Metrics

Upon evaluation, the trainer predicts outputs for the validation dataset, and the computed metrics provide insights into the model's effectiveness. The implementation ensures robust evaluation through a combination of error rates and accuracy metrics, facilitating performance benchmarking and model optimization. By incorporating early stopping, the framework prevents excessive training, mitigating risks of overfitting while maintaining computational efficiency. This structured approach ensures that the OCR model achieves high generalization capability, making it suitable for deployment in real-world text recognition tasks.

### D. *Web Based Integration*

To increase interpretability, a web app is created using vite.js for the frontend that provides an interactive interface for the users. It allows users to either upload an image containing the Devnagari script or write the script directly in a canvas, simulating natural handwriting input. The uploaded files are transmitted to Golang backend server using HTTP, where the images are resized to 384x384 pixels to match the model's requirements. The Golang server communicates with the FastAPI-based server hosting the TrOCR model. The TrOCR model runs script identification on the image and returns the identified Devanagari script in JSON format. The identified script is then visually represented to the user in the web app's UI.

## V. RESULT

### A. *Model Performance and Accuracy*

TABLE I. METRICS

| ***Metrics*** | ***Results*** |
|---|---|
| Character Error Rate (CER) | 3.95% |
| Character Level Accuracy | 96.05% |

As the result of performance evaluation, the model achieves an outstanding performance in the Devnagari Character test set with a minimum CER of 3.95% and an impressive Character-Level Accuracy of 96.05%. These metrics were computed in a uniformly partitioned test set consisting of 467 character images (10% of dataset). The achieved metrics validate the model's robustness to different handwritten character variations.

One thing we noticed is that the model could tell the difference between characters that look almost the same, especially those with small lines or dots added to them Models like CNN usually struggle in such cases—particularly when the handwriting is messy or the letters are close together. But since TrOCR uses a transformer system, it examines the overall structure and context of the image. That likely explains why it performs better

One thing we noticed is that the model could tell the difference between characters that look almost the same, especially those with small lines or dots added to them. . For example, it could accurately distinguish between 'ड' and 'ङ', even though the primary visual difference is a small dot-like mark .Models like CNN usually have trouble in such cases—especially when the handwriting is a bit untidy or two letters are written too close together. But since TrOCR uses a transformer system, it looks at the overall structure of the image and tries to understand the context. This explains why the model performs better even when the writing is not particularly neat or clear.

### B. *Training Efficiency and Convergence*

The preprocessing pipeline consisted of Batch-wise RGB conversion and Normalization which ensured consistency of the input. Additionally, mixed-precision training (fp16) and gradient checkpointing successfully reduced the GPU usage nearly by 30%, that allowed longer training without hardware constraints.

The evaluation loss reduced from 0.9974, at 1% training completion to 0.031 after 6 epochs. Similarly, the training loss reduced from 1.3843 at 1% training completion to 0.0009 after 6 epochs. This demonstrated a stable convergence of the training and evaluation losses.

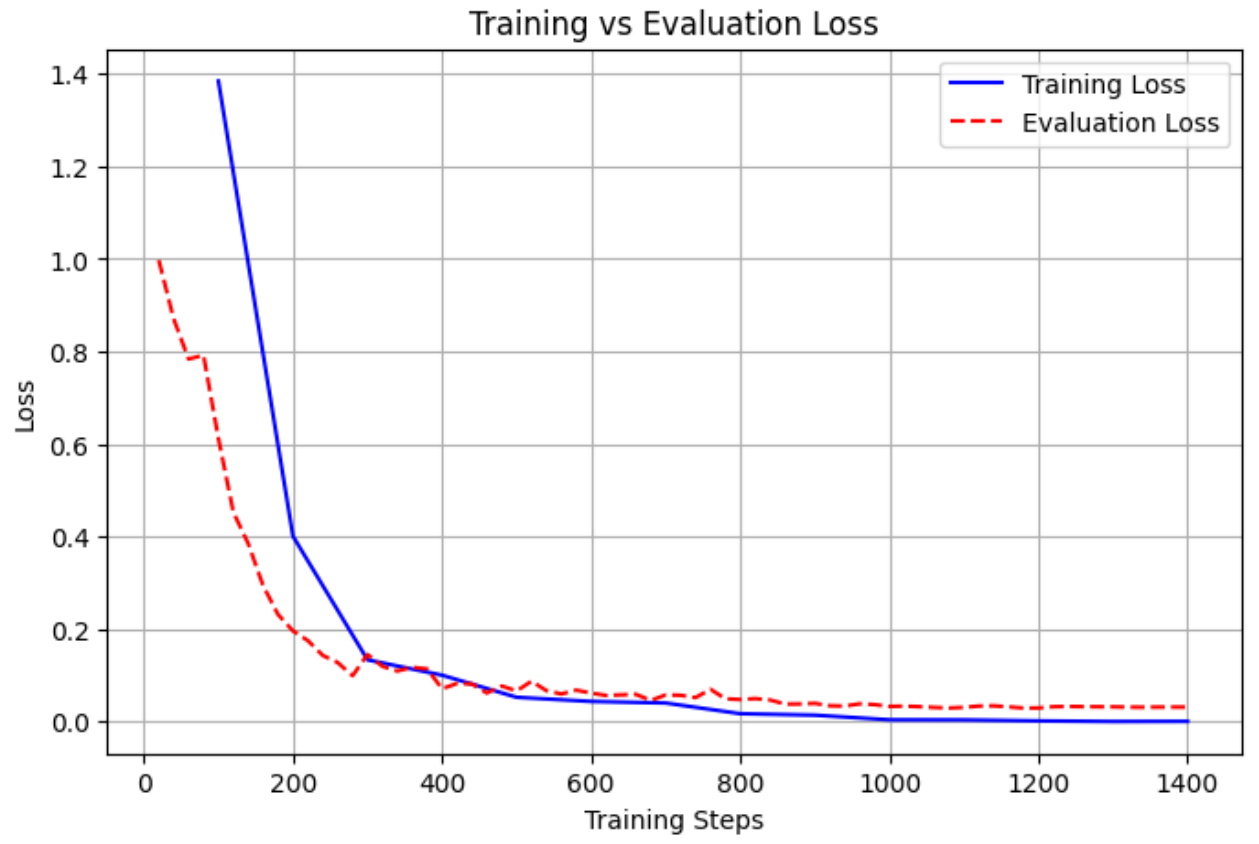

Fig. 6. Training vs Evaluation Losses

### C. *Comparative Analysis with Existing Models*

TABLE II. METRICS COMPARISION

| ***Model*** | ***CER*** | ***Accuracy*** |
|---|---|---|
| ANN [2] | -- | 75.6% |
| Transformer-CNN [4] | 12.7% | 84.3% |
| Template Matching [14] | -- | 92.66% |
| **Proposed TrOCR** | **3.95%** | **96.05%** |

Overall, we can notice that TrOCR's self-attention mechanism, can better handle positional differences and matras in handwritten devnagari script, which allows it to perform better than other conventional approaches in terms of accuracy and CER metrics.

TABLE III. STRENGTHS AND LIMITATIONS ANALYSIS

| *Model* | *Strength* | *Limitation* |
|---|---|---|
| ANN [2] | Works on noisy data. | Poor with conjunct characters. |
| Transformer-CNN [4] | No large annotated data needed. | Lower accuracy than TrOCR. |
| Template Matching [14] | High accuracy for clean samples. | Fails with handwriting variability. |
| **Proposed TrOCR** | Can handle complex conjuncts via self-attention. | Limited to isolated characters |

### *D. Model Deployment and Performance*

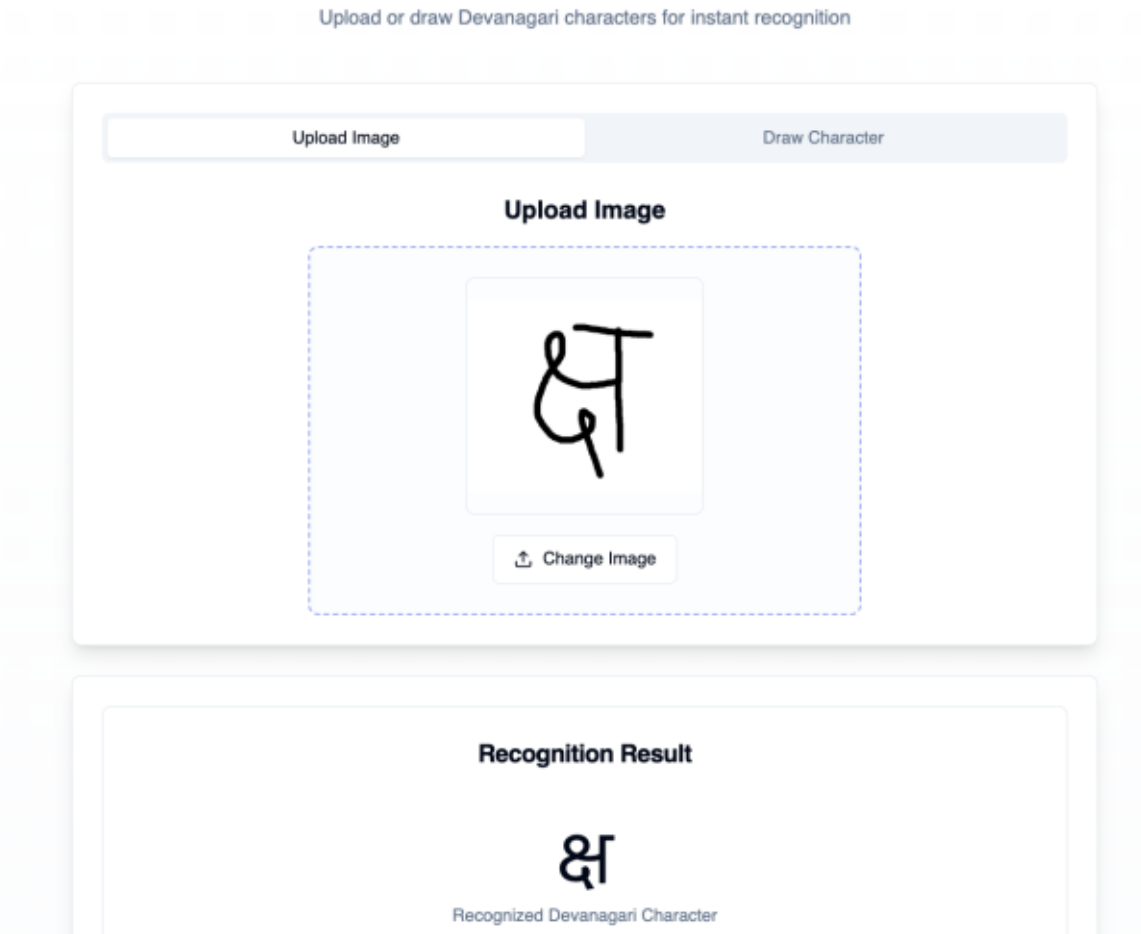

Fig. 7. Image Upload

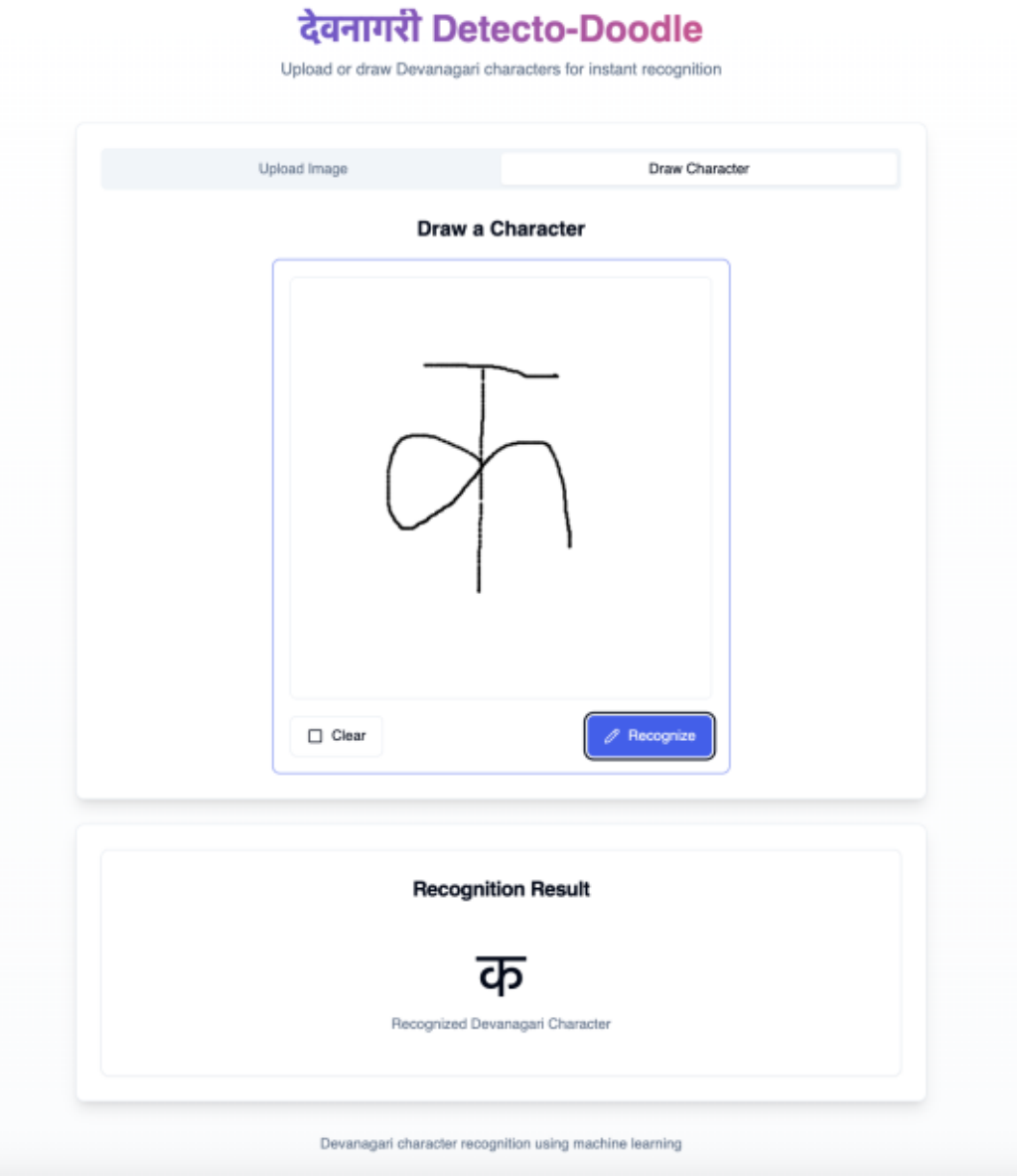

Fig. 8. Canvas Text

The model upon deployment through a web based application, is able to achieve a real-time inference of supplied character data. The standardization of input images to 384x384 pixels preserves the critical image features while minimizing computational overhead. An end-to-end latency of <5s is observed per request. Overall, the Golang-FastAPI backend is able to serve numerous concurrent user requests with quick responses and minimal latency.

## VI. FUTURE WORK

Currently the existing dataset focuses on isolated Devnagari characters which limits the model's ability to recognize words and sentences. So, the dataset can be expanded to Devnagari handwritten words and sentences. These word and sentence images may be derived from ancient handwritten Sanskrit scripts, modern writings of writers across various age groups and regions in the Indian subcontinent which ensures sufficient diversity in the dataset. Using this large scale and diverse dataset, the model can be further fine-tuned with high-end hardware and GPUs making it able to recognize diverse Devnagari scripts across multiple regions and timeline.

## VII. CONCLUSION

This study successfully fine-tunes the TrOCR model for handwritten Devnagari Handwritten Character Recognition with an impressive CER of 3.95% and Character Accuracy of 96.05% using a diverse set of Devnagari character dataset consisting of numerals, vowels and consonants.

The OCR model was then integrated into a scalable web application Golang, FastAPI and Vite.js framework that achieved a minimal latency of <5s per request. The system proposed in this study bridges critical gap from Devnagari handwritten characters to digitized Devnagari characters. Future work will further expand this framework to recognize and analyze multiple words and sentences from Devnagari scripts. Overall, this study creates a solid foundation for a scalable solution that may digitize variety of handwritten Devnagari scripts from writers across multiple regions, age groups and timelines that can have high significance in educational, historical and administrative context.